\documentclass[a4paper, 12pt]{article}
\usepackage{arxiv}

\usepackage[utf8]{inputenc}
\usepackage[T1]{fontenc}
\usepackage{hyperref}       
\usepackage{url}           
\usepackage{booktabs}     
\usepackage{amsfonts}     
\usepackage{nicefrac}      
\usepackage{microtype}  
\usepackage{lipsum}	
\usepackage{graphicx}
\usepackage{natbib}
\usepackage{doi}
\usepackage{amsmath,graphicx}

\title{HRHD-HK: A Benchmark Dataset of High-Rise and High-Density Urban Scenes for 3D Semantic Segmentation of Photogrammetric Point Clouds}

\author{Maosu Li, Yijie Wu, Anthony G.O. Yeh, Fan Xue \thanks{\textbf{Citation}: Li, M., Wu, Y., Yeh, A. G. O., \& Xue, F. (2023). HRHD-HK: A benchmark dataset of high-rise and high-density urban scenes for 3D semantic segmentation of photogrammetric point cloud. \emph{Proceedings of 2023 IEEE International Conference on Image Processing Challenges and Workshops}, 3714-3718. IEEE. \url{https://doi.org/10.1109/ICIPC59416.2023.10328383} }\\
	Faculty of Architecture, The University of Hong Kong, Hong Kong SAR, China\\
	\texttt{\{maosulee, yijiewu\}@connect.hku.hk, \{hdxugoy, xuef\}@hku.hk} \\
}

\renewcommand{\shorttitle}{HRHD-HK: A Benchmark Dataset}

\hypersetup{
pdftitle={HRHD-HK: A Benchmark Dataset of High-Rise and High-Density Urban Scenes for 3D Semantic Segmentation of Photogrammetric Point Clouds},
pdfsubject={cs.CV, cs.DB, eess.IV},
pdfauthor={Maosu Li, Yijie Wu, Anthony G.O. Yeh, Fan Xue},
pdfkeywords={Benchmark dataset, photogrammetric point clouds, 3D semantic segmentation, high-rise high-density city, deep learning},
}

\begin{document}

\date{}

\maketitle
\begin{abstract}
Many existing 3D semantic segmentation methods, deep learning in computer vision notably, claimed to achieve desired results on urban point clouds. Thus, it is significant to assess these methods quantitatively in diversified real-world urban scenes, encompassing high-rise, low-rise, high-density, and low-density urban areas. However, existing public benchmark datasets primarily represent low-rise scenes from European cities and cannot assess the methods comprehensively. This paper presents a benchmark dataset of high-rise urban point clouds, namely High-Rise, High-Density urban scenes of Hong Kong (HRHD-HK). HRHD-HK arranged in 150 tiles contains 273 million colorful photogrammetric 3D points from diverse urban settings. The semantic labels of HRHD-HK include building, vegetation, road, waterbody, facility, terrain, and vehicle. To our best knowledge, HRHD-HK is the first photogrammetric dataset that focuses on HRHD urban areas. This paper also comprehensively evaluates eight popular semantic segmentation methods on the HRHD-HK dataset. Experimental results confirmed plenty of room for enhancing the current 3D semantic segmentation of point clouds, especially for city objects with small volumes. Our dataset is publicly available at \url{https://doi.org/10.25442/hku.23701866.v2}.
\end{abstract}
\section{Introduction}
\label{sec:1}

\begin{figure}[tb]
\begin{minipage}[tb]{1.0\linewidth}
  \centering
  \includegraphics[width=\linewidth]{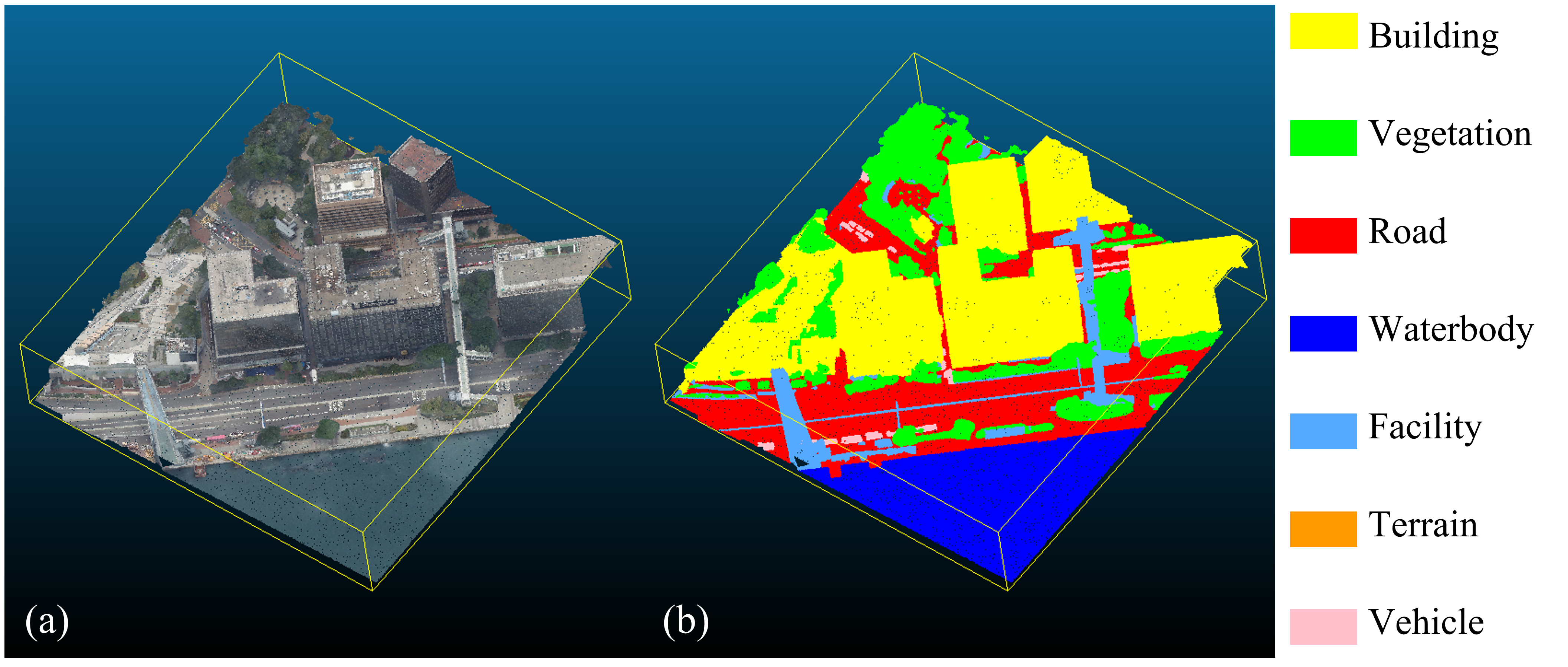}
\end{minipage}
\caption{Example of 3D semantic segmentation in the HRHD-HK dataset. (a) Unstrucured 3D data; (b) semantic labels.}
\label{fig:1}
\end{figure}

Fully semantic-enriched 3D point clouds play a significant role in smart city applications, such as robotics, autonomous driving and navigation, and urban analytics~\cite{guo2020deep,li2023bi}. 3D semantic segmentation of point clouds is the process that assigns each point a semantic label, such as building, vegetation, road, and waterbody, as shown in Figure~\ref{fig:1}, in order to enable vehicles and robots to comprehend city objects' functions and morphology. A variety of 3D semantic segmentation methods, deep learning notably, have been popularized in computer vision, photogrammetry, and remote sensing fields~\cite{guo2020deep}. Semantic enrichment of point clouds highly relies on these automatic methods, because the sheer size and diversity of the represented city objects make the manual judgment high-cost and inefficient.

Assessing the 3D semantic segmentation methods quantitatively in real-world urban scenes is important. Generally, methods should be evaluated comprehensively on worldwide datasets of diverse urban settings, including high-rise, low-rise, high-density, and low-density urban scenes. However, as listed in Table~\ref{table:1}, existing public benchmark datasets, such as Swiss3DCities~\cite{can2021semantic} and DublinCity~\cite{Zolanvari2019}, primarily represent low-rise scenes from European cities. In contrast, benchmark datasets of high-rise, high-density (HRHD) urban scenes, e.g., in Hong Kong (HK), New York, and Tokyo, are absent. Consequently, the absence of datasets of HRHD urban scenes fundamentally undermines the comprehensiveness and accuracy of the assessment of 3D semantic segmentation methods. E.g., the same 3D semantic segmentation model can show inconsistent performances on different urban morphological point clouds~\cite{hu2022sensaturban}.


\begin{table*}[tb]
\centering
\caption{List of benchmark datasets for 3D semantic segmentation of photogrammetric point clouds}\label{table:1}
\footnotesize
\resizebox{\textwidth}{!}{
\begin{tabular}{lllllllllll}
\toprule
By & Dataset & Location & Avg. bldg. & Avg. bldg. & Area & Points & No. of & Color? & HRHD? & Source  \\ 
 & &  & height (m) & cov. ratio & (km$^2$) & (Mil.) &  classes & &  &   \\ 
\midrule
\cite{AHN19} & AHN3  & Netherland & 15.66$^\dag$ & 0.44$^\dag$ & 41,543 & 4E+5$^\ddag$ & 4 & No  & No & LiDAR \\
\cite{Zolanvari2019} & Dublin City & Ireland & 22.56 & 0.64 & 2 & 260 & 13 & No  & No & LiDAR \\
\cite{varney2020dales} & DALES & Canada & 8.83 & 0.39 & 10 & 505 & 8 & No & No & LiDAR \\
\cite{li2020campus3d} & Campus3D & Singapore & 23.63 & 0.26 & 1.58 & 937 & 24 & Yes & No & Photogrammetry \\
\cite{can2021semantic} & Swiss3DCities &  Switzerland & 14.85 & 0.39 & 2.70 & 226 & 5 & Yes & No &  Photogrammetry \\
\cite{kolle2021hessigheim} & Hessigheim3D & Germany & 4.86 & 0.30 & 0.19 & 126 & 11 & Yes & No & LiDAR \\
\cite{hu2022sensaturban} & SensatUrban & United Kingdom & 7.67 & 0.33 & 7.64 & 2,847 & 13 & Yes & No &  Photogrammetry \\
\midrule
  &  {\bf Our HRHD-HK} & Hong Kong & 38.50 & 0.67$^\dag$ & 9.38 &  273 & 7 & Yes & Yes & Photogrammetry \\ 
\bottomrule
\multicolumn{11}{l}{$\dag$ indicates computed values in urban central areas; $\ddag$ denotes an estimated total number of points.}
\end{tabular}
}
	\vspace{-2em}
\end{table*}

This paper presents HRHD-HK, a benchmark dataset of HRHD urban scenes for 3D semantic segmentation of photogrammetric point clouds. The semantic labels of HRHD-HK include building, vegetation, road, waterbody, facility, terrain, and vehicle. Point clouds of HRHD-HK were collected in HK with two features, i.e., color and coordinates. HRHD-HK arranged in 150 tiles, contains approximately 273 million points, covering 9.375 km$^2$. HRHD-HK aims to supplement the existing benchmark datasets by incorporating Asian HRHD urban scenes as well as subtropical natural landscapes, such as sea, vegetation, and mountains.

The contribution of this paper is two-fold. First, it presents the first public benchmark dataset of HRHD urban scenes for 3D semantic segmentation of photogrammetric point clouds. The HRHD urban morphologies of the HRHD-HK dataset supplement the current benchmark datasets. Secondly, we provide a comprehensive evaluation of eight popular 3D semantic segmentation methods on HRHD-HK. Experimental results confirmed plenty of room for enhancing the current 3D semantic segmentation of point clouds in HRHD urban areas.

\section{HRHD-HK: Building high-rise high-density photogrammetry dataset with urban semantics}
\label{sec:2}
HK is a city with typical HRHD morphologies across the world. E.g., there exist 2,522 buildings over 100 m with 309 skyscrapers above 150 m. The maximum building coverage ratio of blocks exceeds 0.70. Compared to existing benchmark datasets with low-rise building blocks as shown in Table~\ref{table:1}, the average height of buildings of HRHD-HK is 38.50 m, where 163 buildings are higher than 100 m. The average building coverage ratio of blocks in downtown areas reaches up to 0.67. Specifically, we created the HRHD-HK dataset through four steps.

\begin{figure}[tb]
\begin{minipage}[tb]{1.0\linewidth}
  \centering
  \includegraphics[width=\linewidth]{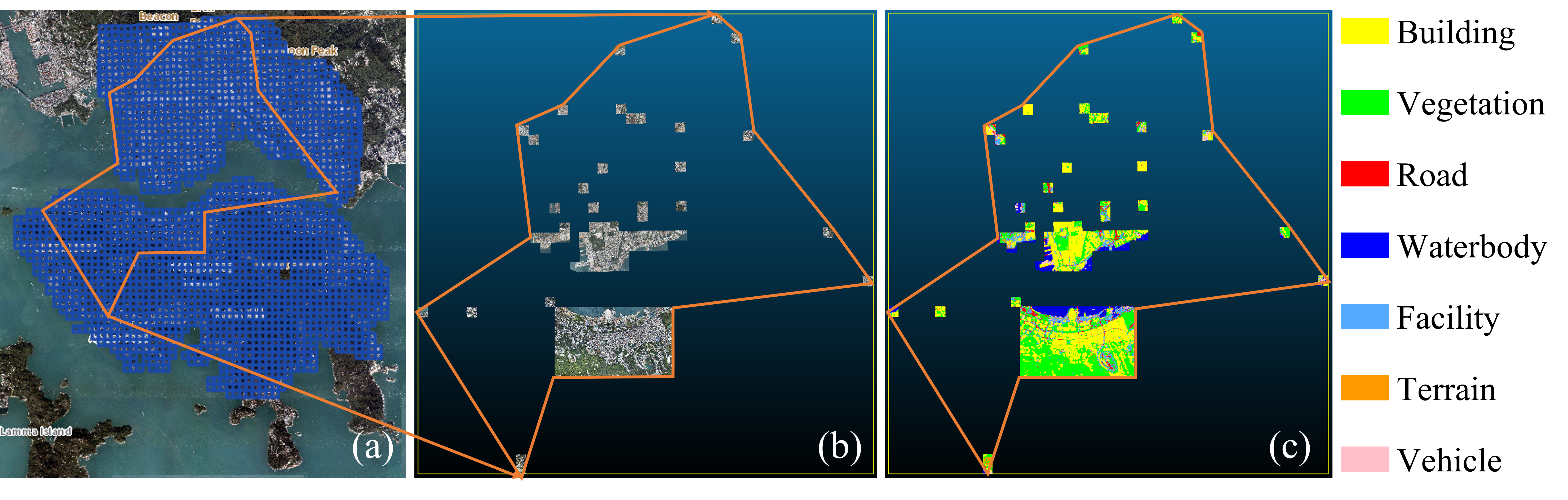}
\end{minipage}
\caption{Creation process of HRHD-HK. (a) Spatial range of selected photo-realistic mesh models; (b) selected 150 tiles of photo-realistic mesh models; (c) sampled point clouds with semantic labels.}
\label{fig:2}
\end{figure}

\begin{figure}[tb]
\begin{minipage}[tb]{1.0\linewidth}
  \centering
  \includegraphics[width=\linewidth]{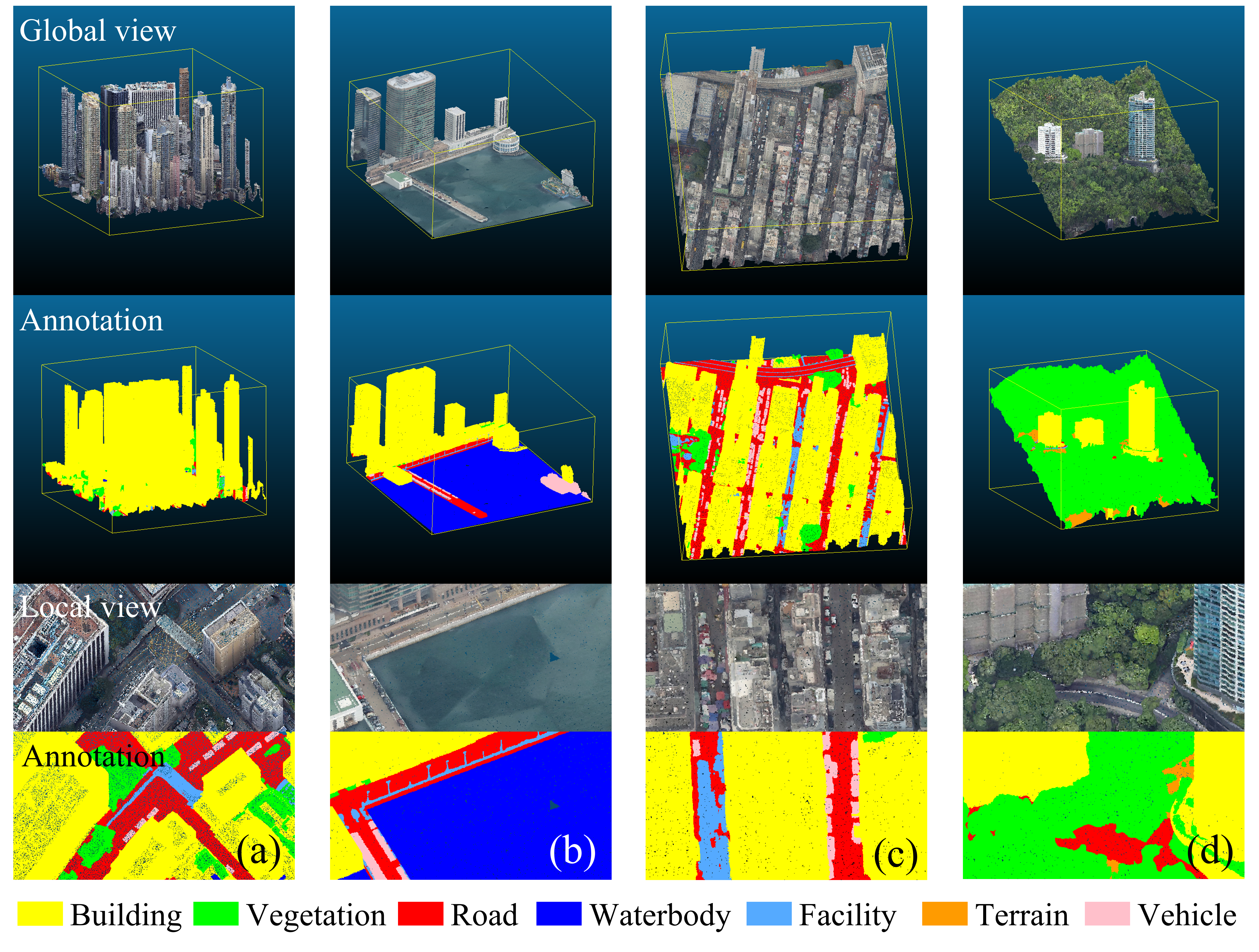}
\end{minipage}
\caption{Example morphologies of point clouds in HRHD-HK. (a) High-rise, high-density building blocks; (b) harbor; (c) high-density low-rise flats; (d) green hills.}
\label{fig:3}
\end{figure}

Step 1: Data acquisition. The point cloud dataset was generated from the photo-realistic mesh models~\cite{HKPlanD2019} provided by the HK Planning Department. Figure~\ref{fig:2}a shows the original photo-realistic mesh models arranged in 2,150 tiles, covering nearly the whole HK Island and the main urban area of Kowloon Peninsula. Figure~\ref{fig:2}b shows the selected 150 square tiles (9.375 km$^2$ in total) of photo-realistic mesh models from concentrated and separated urban areas. Specifically, one of the most densely developed downtown areas on both sides of Victoria Harbor of HK was selected to completely represent the HRHD urban morphologies. Figure~\ref{fig:3}a shows typical HRHD building blocks in HK. Thereafter, separated tiles were also selected to incorporate more other morphologies in HK, such as clusters of low-rise flats and green hills as shown in Figures~\ref{fig:3}c and \ref{fig:3}d.

Step 2: Data curation. We first manually removed the incorrectly reconstructed triangle faces from the photo-realistic mesh models. E.g., trivial triangle faces separated from the main body of the tile were removed as outliers. At the sampling density of 10 points per square meter, the 150 tiles of point clouds sampled from mesh models contain about 273 million points. Last, we geo-registered all points in the HK 1980 Grid (EPSG:2326), where the unit of coordinates \textit{xyz} is meter.

Step 3: Semantic annotation. Given the HRHD-HK dataset with more than 273 million points and the crowded spatial distribution of landscape elements, we outsourced the annotation task to professional annotators for high-quality ground truth. To ensure the annotation quality, we manually checked each tile of HRHD-HK. Point clouds with incorrect annotations were returned for modification until there existed no observable errors. Point clouds of HRHD-HK were annotated into seven semantic categories, as shown in Figure~\ref{fig:2}c and Table~\ref{table:2}. The seven semantic labels are building, vegetation, road, waterbody, facility, terrain, and vehicle, which represent the most common city objects in HK. E.g., building denotes constructions with both walls and roofs in HRHD-HK, and points representing non-building objects, e.g., fences, canopies, containers, footbridges, and billboards were labeled as facility. Figure~\ref{fig:3} shows example tiles of point clouds with diverse urban morphologies, i.e., HRHD building blocks, harbor, high-density but low-rise flats, and green hills.

\begin{table}[tb]
	\centering
	\caption{List of seven semantic labels with example objects}\label{table:2}
	\footnotesize
		\begin{tabular}{cll}
			\toprule
			No & Label & Description  \\
			\midrule
			1 & Building    & Construction with walls and a roof. \\
			2 & Vegetation  & Tree, bush, grass, etc. \\ 
			3 & Road        & All types of uncovered constructed path, e.g., road, \\
			& & street, walkway, and flyover. \\
			4 & Waterbody   & Sea, lake, swimming pool, etc.  \\
			5 & Facility    & Fence, canopy, container, footbridge, billboard, etc.\\
			6 & Terrain     & Unconstructed land surface, e.g., bare earth and \\
			& & constructed slope made of concrete.  \\
			7 & Vehicle     & Car, ship, vessel, etc. \\
			\bottomrule
		\end{tabular}
		\vspace{-2em}
\end{table}

\begin{figure}[tb]
	\begin{minipage}[tb]{1.0\linewidth}
		\centering
		\includegraphics[width=\linewidth]{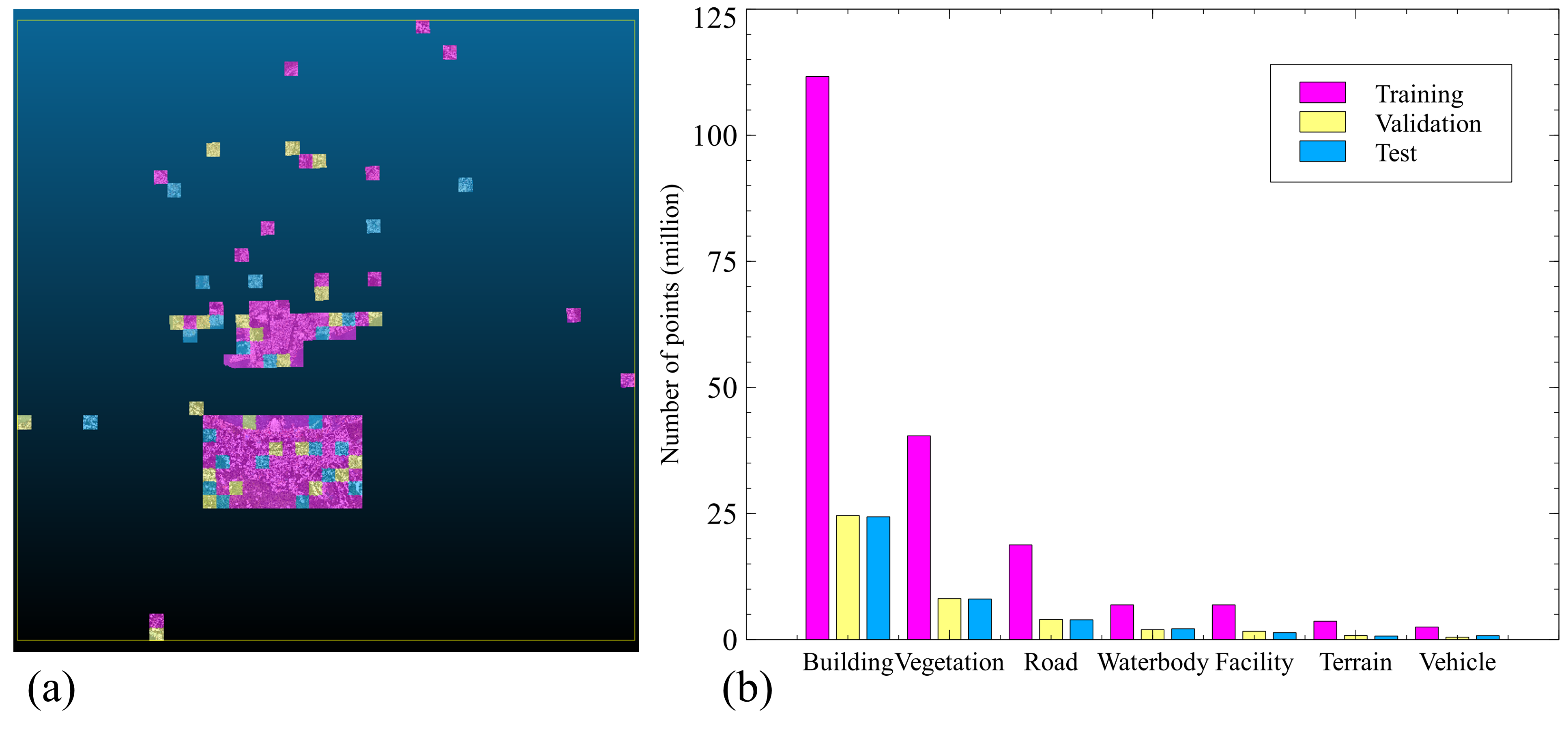}
	\end{minipage}
	\caption{Distribution of training, validation, and test sets. (a) Spatial and (b) quantity distributions.}
	\label{fig:4}
\end{figure}

Step 4: Training-validation-test split. We selected 104 tiles of HRHD-HK as the training set, another 23 tiles of HRHD-HK as the validation set, and the last 23 tiles as the test set. Due to the imbalanced spatial distribution of city objects, we manually arranged the training, validation, and test sets (see Figure~\ref{fig:4}a) to balance the quantity distribution of different landscape elements, especially for city objects represented by small numbers of points (e.g., waterbody, facility, terrain, and vehicle) in validation and test sets. Figure~\ref{fig:4}b shows the numbers of points representing seven landscape elements for validation and test sets were all non-zero and almost equal.

\section{Evaluation of 3D semantic segmentation methods on HRHD-HK}
\label{sec:3}

\textbf{Segmentation methods}. We evaluated eight typical 3D semantic segmentation methods from projection-based, voxel-based, and point-wise semantic segmentation architectures on HRHD-HK. Specifically, the eight methods were a 3D voxel convolution network SparseConvUnet~\cite{graham20183d}, a 2D projection method BEV-Seg3D-Net~\cite{zou2021efficient}, a graph method SPGraph~\cite{landrieu2018large}, a Kernel point convolution method KPConv~\cite{thomas2019kpconv}, three multi-layer perceptron methods including PointNet~\cite{qi2017pointnet}, PointNet++~\cite{qi2017pointnet++}, and RandLA-Net~\cite{hu2020randla}, and a transformer named StratifiedTransformer~\cite{lai2022stratified}.

\textbf{Evaluation metrics}. Three commonly used indicators including Overall Accuracy (OA), mean class Accuracy (mAcc), and mean Intersection over Union (mIoU) were applied to evaluate the performance of the eight selected methods on HRHD-HK. OA reports the percentage of total points which are correctly classified, whereas mAcc represents the average percentage of points that are correctly classified in each class. mIoU is the average of the IoUs, which indicates the average magnitude of the detection confusion of each semantic label.

\textbf{Training environment}. The training environment was set up as follows. The experiments were implemented on a high-performance computing cluster with 7 servers, each of which owns dual Intel Xeon 6226R (16 core) CPUs, 384GB RAM, 4 $\times$ NVIDIA V100 (32GB) SXM2 GPUs, and a CentOS 8 system. Each deep learning model was trained on assigned 16-core CPUs, 64GB RAM, and one NVIDIA V100 (32GB) SXM2 GPU card. All eight models were trained with the environment of PyTorch (ver. 1.8) and Python (ver. 3.7). We universally used 0.15-meter downsampling to preprocess the point clouds. Hypermeters of eight semantic segmentation methods were fine-tuned to achieve the best results we could acquire.

\begin{table*}[tb]
	\centering
	\caption{OA, mAcc, mIoU and per-class IoUs of selected methods (best value in each column in bold)}\label{table:3}
	\footnotesize
		\begin{tabular}{@{}ll@{}c@{\;}c@{}cc@{}c@{}c@{}c@{}c@{}c@{}c@{}}
			\toprule
			Group & DL method & \multicolumn{3}{c}{Overall metric (\%)} & \multicolumn{7}{c}{Per-class IoU (\%)} \\
		\cmidrule(lr){3-5}\cmidrule(lr){6-12}
			 &  & OA & mAcc & mIoU & Building & Vegetation  & Road  & Waterbody & Faility & Terrain & Vehicle  \\
			\midrule
			Voxel & SparseConvUnet       \cite{graham20183d} & 88.71 & 70.24 & 58.46 & 90.71 & 88.31 & 57.99 & 93.15 & 24.60 & 25.09 & 29.38 \\
			2D proj. & BEV-Seg3D-Net        \cite{zou2021efficient} & 89.18 & 73.25 & 61.14 & 90.75 & 88.34 & 54.88 & 94.53 & 25.17 & 31.12 & 43.20 \\
			Graph & SPGraph              \cite{landrieu2018large} & 85.32 & 58.53 & 49.86 & 86.65 & 78.19 & 50.28 & 91.93 & 14.20 & 14.71 & 13.08 \\
			Kernel & KPConv               \cite{thomas2019kpconv} & 91.23 & 71.53 & 63.81 & 92.39 & 88.56 & 62.89 & 91.96 & 27.19 & 34.33 & 49.38\\
			MLP & PointNet            \cite{qi2017pointnet}  & 77.49 & 61.98 & 47.50 & 81.09 & 58.39 & 51.07 & 92.34 & 11.84 & 19.74 & 18.07 \\
			 & PointNet++           \cite{qi2017pointnet++} & 79.85 & 66.95 & 52.52 & 77.43 & 58.55 & 58.52 & 95.16 & 21.40 & 29.88 & 26.98 \\ 
			 & RandLA-Net           \cite{hu2020randla} & 90.39 & \textbf{78.81} & 64.76 & 91.29 & 90.39 & 63.59 & 94.21 & 32.24 & 36.45 & 45.13 \\
			Trans. & StratifiedTransformer \cite{lai2022stratified} & \textbf{92.30} & 76.99 & \textbf{68.08} & \textbf{93.17} & \textbf{91.99} & \textbf{67.35} & \textbf{95.61} & \textbf{35.31} & \textbf{38.19} & \textbf{54.91} \\ 
			\bottomrule
		\end{tabular}
	\vspace{-2em}
\end{table*}

\textbf{Results}. Table~\ref{table:3} lists the evaluation results of the eight methods. Because of the multi-scale receptive size and the attention mechanism, the most up-to-date method, StratifiedTransformer published last year achieved the best performance of OA and mIoU at 92.30\% and 68.08\%, respectively, whereas RandLA-Net achieved the highest value of mAcc at 78.81\%. By contrast, PointNet received the lowest OA and mIoU. Although the detection of city objects such as buildings, vegetation, and waterbody achieved relatively high-level performance with the highest per-class IoUs above 91.99\%, city objects such as roads, terrain, vehicles, and facilities were still poorly segmented. E.g., the highest IoUs of road, vehicle, terrain, and facility were 67.35\%, 54.91\%, 38.19\%, and 35.31\%, respectively. 

\begin{figure}[h!]
  \centering
  \includegraphics[width=0.82\linewidth]{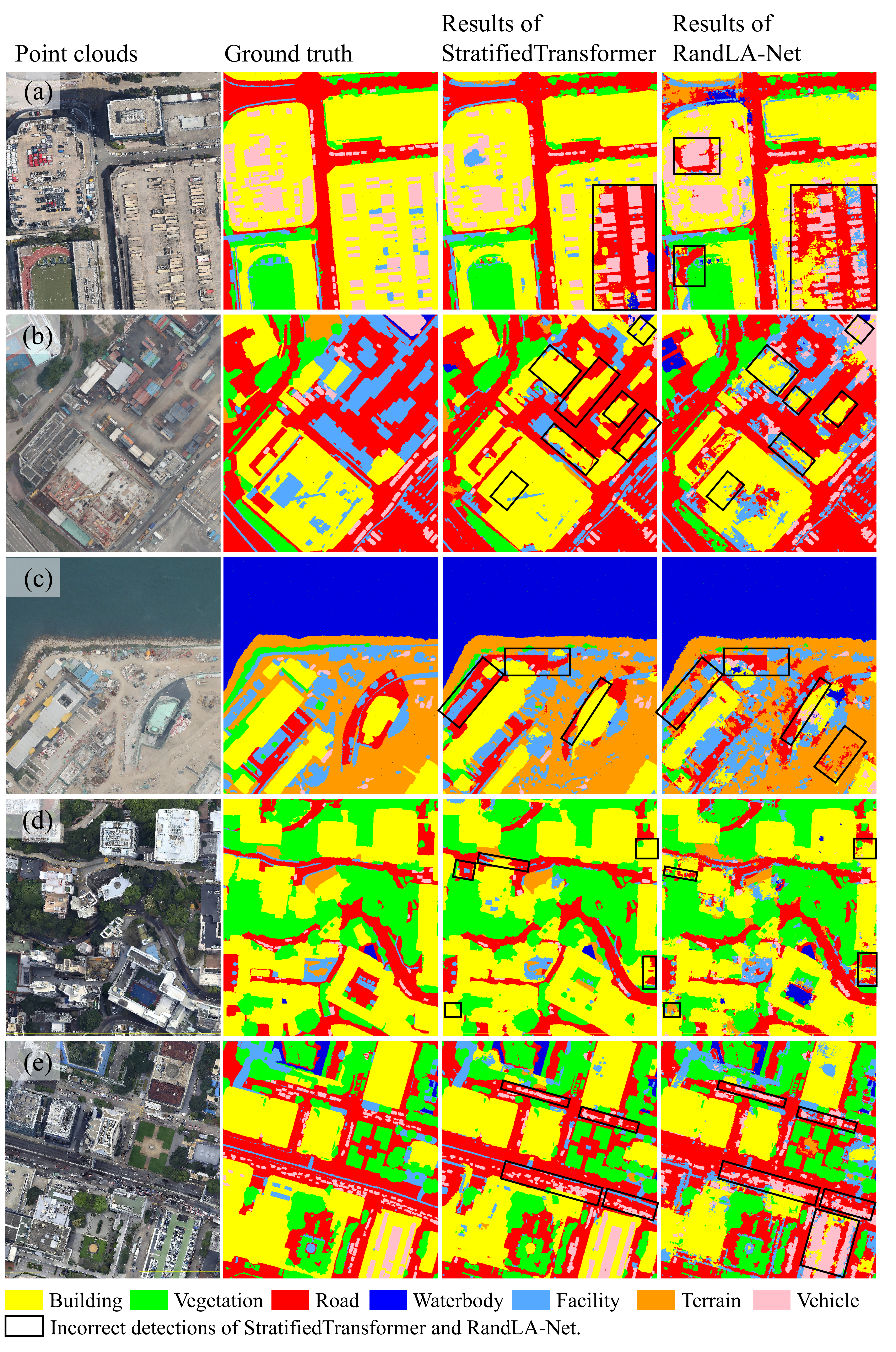}
\caption{Typical errors of StratifiedTransformer and RandLA-Net on HRHD-HK. (a) Confusion between large-size building roofs and roads; (b)-(d) confusions between facility and building, roads and terrain, and roads and building podiums; (e) poor detection of vehicles.}
\label{fig:5}
\end{figure}

\textbf{Discussion}. Figure~\ref{fig:5} shows typical confusions through the inference results of StratifiedTransformer and RandLA-Net as examples. First, the large sizes of certain parts of city objects lead to incorrect detections. E.g., large flat surfaces such as building roofs were incorrectly detected into roads as shown in Figure~\ref{fig:5}a, because large building roofs are similar to pavement and roads.

Thereafter, small city objects with similar appearances are difficult to be distinguished. E.g., Figure~\ref{fig:5}b shows facilities like containers at the dock could be easily detected as buildings. Figure~\ref{fig:5}c shows terrain like bare earth on the ground level was confused with constructed roads. There exist difficulties in distinguishing narrow roads from adjacent building podiums in the mountainous and multilevel urban environment as shown in Figure~\ref{fig:5}d. 

Last, the extremely high size ratio between large city objects such as building blocks and greenery, and small city objects such as vehicles and facilities poses challenges to balancing detection performance. Figure~\ref{fig:5}e shows most greenery and buildings were detected whereas the boundaries of vehicles were difficult to be distinguished from the roads and building roofs, especially for RandLA-Net.

Overall, there still exists room for selected 3D semantic segmentation methods to achieve satisfactory performance in HRHD-HK, especially for city objects with small volumes such as roads, vehicles, terrain, and facilities in the HRHD urban context.

\section{Conclusion}
A variety of urban point cloud benchmark datasets is significant in training, examining, and advancing 3D semantic segmentation methods for diversified urban scenes across the world. However, current benchmark datasets of photogrammetric point clouds primarily represent low-rise urban morphologies, especially in European cities. This paper presents the first high-rise, high-density (HRHD) urban benchmark dataset, HRHD urban scenes of Hong Kong (HRHD-HK) to supplement the current benchmark dataset hub.

The proposed HRHD-HK covers 9.375 km$^2$ of urban areas of HK with 273 million color points. HRHD-HK includes seven semantic labels, i.e., building, vegetation, road, waterbody, facility, terrain, and vehicle. We tested eight 3D semantic segmentation methods, all of which had mIoUs less than 68.08\%. Particularly, there exists room for improvement in detecting city objects with small volumes such as roads, vehicles, terrain, and facilities in the HRHD urban context. We make HRHD-HK publicly available for researchers to benchmark deep learning methods and advance their generalization in HRHD cities. Our future work includes embedding publicly available geospatial information to extend the dimension of model training for more accurate 3D semantic segmentation.

\section*{Acknowledgement}
This study was supported in part by the Hong Kong Research Grant Council (RGC) (27200520) and Department of Science and Technology of Guangdong Province (GDST) (2020B1212030009, 2023A1515010757).

\bibliographystyle{unsrtnat}
\bibliography{refs}

\end{document}